\documentclass[11pt,a4paper]{article}
\usepackage[hyperref]{acl2021}
\usepackage{times}
\usepackage{latexsym}
\usepackage{graphicx}
\usepackage{subfigure}
\usepackage{url}
\usepackage{booktabs}
\usepackage{multirow}
\usepackage{amsmath}
\usepackage{amsfonts}
\usepackage{diagbox}

\usepackage{microtype}

\aclfinalcopy 

\title{\emph{Conversations Are Not Flat}:\\ Modeling the Dynamic Information Flow across Dialogue Utterances}

\author{
    Zekang Li$^1$$^2$, Jinchao Zhang$^3$, Zhengcong Fei$^1$$^2$, Yang Feng$^1$$^2$\thanks{\ \ Joint work with Pattern Recognition Center, WeChat AI, Tencent Inc. Yang Feng is the corresponding author. Work was done when Zekang Li and Zhengcong Fei were intern at WeChat AI.}, \bf{Jie Zhou}$^3$ \\
  $^{1}$ Key Laboratory of Intelligent Information Processing \\
  Institute of Computing Technology, Chinese Academy of Sciences (ICT/CAS) \\
  $^{2}$ University of Chinese Academy of Sciences \\
  $^{3}$ Pattern Recognition Center, WeChat AI, Tencent Inc, China \\
  {\tt \{\href{mailto:lizekang19g@ict.ac.cn}{lizekang19g},\href{mailto:feizhengcong@ict.ac.cn}{feizhengcong},\href{mailto:fengyang@ict.ac.cn}{fengyang}\}@ict.ac.cn} \\
  {\tt \{\href{mailto:dayerzhang@tencent.com}{dayerzhang},\href{mailto:withtomzhou@tencent.com}{withtomzhou}\}@tencent.com}
  \\}

\date{}

\begin{document}
\maketitle

\begin{abstract}
Nowadays, open-domain dialogue models can generate acceptable responses according to the historical context based on the large-scale pre-trained language models.
However, they generally concatenate the dialogue history directly as the model input to predict the response, which we named as the \emph{flat pattern} and ignores the dynamic information flow across dialogue utterances. 
In this work, we propose the \textbf{DialoFlow} model, in which we introduce a dynamic flow mechanism to model the context flow,
and design three training objectives to capture the information dynamics across dialogue utterances by addressing the semantic influence brought about by each utterance in large-scale pre-training.
Experiments on the multi-reference Reddit Dataset and DailyDialog Dataset demonstrate that our DialoFlow significantly outperforms the DialoGPT on the dialogue generation task.
Besides, we propose the \textbf{Flow score}, an effective automatic metric for evaluating interactive human-bot conversation quality based on the pre-trained DialoFlow, which presents high chatbot-level correlation ($r=0.9$) with human ratings among 11 chatbots. 
Code and pre-trained models will be public. \footnote{\url{https://github.com/ictnlp/DialoFlow}}
\end{abstract}

\section{Introduction}

Recent intelligent open-domain chatbots \cite{adiwardana2020towards,bao2020plato,smith2020can} have made substantial progress thanks to the rapid development of the large-scale pre-training approaches \cite{devlin2019bert,radford2019language,brown2020language} and the large amount of conversational data \cite{dinan2018wizard,baumgartner2020pushshift,smith2020can}. 
However, effectively modeling the dialogue history in large-scale dialogue pre-training is still challenging. 

Most of the previous work on dialogue history modeling mainly fall into two groups. One group of works generally concatenate the dialogue history as the model input and predict the response \cite{zhang2019dialogpt,smith2020can,bao2020plato}, named as \emph{flat pattern}, which is commonly adopted in the large-scale pre-training. However, \citet{sankar2019neural} demonstrate that flat concatenation is likely to ignore the conversational dynamics across utterances in the dialogue history. Another group of works employ \emph{hierarchical modeling} to encode the dialogue history \cite{DBLP:conf/aaai/SerbanSBCP16,DBLP:conf/acl/ShanLZMFNZ20,gu2020dialogbert}, in which the utterances are separately encoded and then fed into an utterance-level encoder. These approaches lack the history information when encoding each individual utterance, while the history information is essential for understanding dialogue utterances. Thus, all the aforementioned methods are deficient in modeling the dynamic information in the dialogue history.

\begin{figure}[t]
    \centering
    \includegraphics[width=0.45 \textwidth]{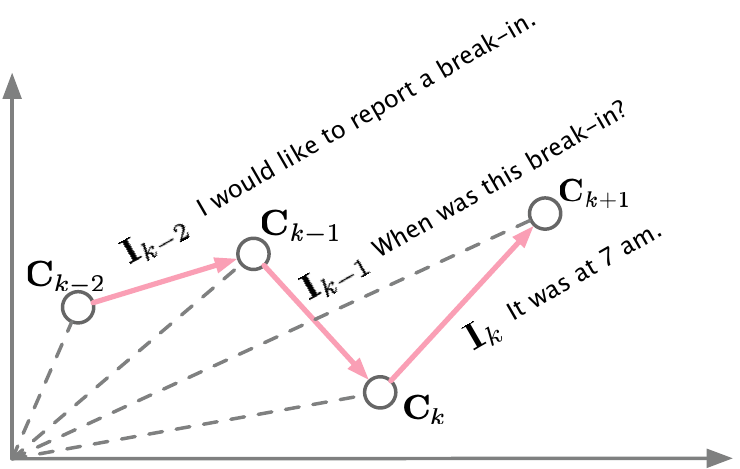}
    \caption{Illustration of the dynamic information flow in the semantic space. $\mathbf{C}_k$ (gray dotted line) denotes the dense representation of dialogue history which we named as context. $\mathbf{I}_k$ (pink solid line) denotes the semantic influence brought about by the $k$-th utterance, which is the difference between $\mathbf{C}_{k}$ and $\mathbf{C}_{k+1}$.}
    \label{fig:flow}
\end{figure}

In this work, inspired by the human cognitive process that humans always consider the goal or influence of the next response before they continue the conversation \cite{brown2015processes}, we propose the DialoFlow to model the dynamic information flow in the dialogue history by addressing the semantic influence brought about by each utterance. As shown in Figure \ref{fig:flow}, we define the dense representation of the dialogue history at different utterances as the \emph{context}s (gray dot line) and the context transformation as the \emph{semantic influence} brought by each utterance. In particular, our DialoFlow constructs the process of the utterance-level history context flow. Correspondingly, the semantic influence of each utterance can be measured by the difference between two adjacent contexts, which will be further used to guide the current response generation.

Practically, we first employ a transformer to encode the whole conversation to get the dense context representation. Then we design a uni-directional Flow module to capture the context flow on the utterance level, and design three training objectives to model the context flow and measure the semantic influence brought about by each utterance: 
1) \emph{Context Flow Modeling}, which aims to capture the context flow schema.  2) \emph{Semantic Influence Modeling}, which targets to measure the predicted semantic influence.  3) \emph{Response Generation Modeling}, which is to generate the response under the guidance of the predicted semantic influence.
Furthermore, to demonstrate the effect of modeling dynamic information flow in the dialogue understanding, we propose the \textbf{Flow score} based on the DialoFlow, an automatic reference-free evaluation metric for interactive dialogue evaluation by measuring the semantic influence perplexity. 

We pre-train the proposed DialoFlow on the large-scale Reddit comments and conduct experiments on dialogue generation and interactive dialogue quality evaluation. For dialogue generation, DialoFlow achieves significant improvements on the Reddit multi-reference dataset and the DailyDialog dataset compared to the baseline DialoGPT \cite{zhang2019dialogpt}. 
For interactive dialogue quality evaluation, our proposed Flow score obtains an impressively high chatbot-level correlation ($r=0.9$) with human ratings on 2200 human-bot dialogues from 11 chatbots.

Our contributions are summarized as follows:
\begin{itemize}
    \item We propose the \textbf{DialoFlow}, a new paradigm to construct the dynamic information flow in the dialogue history by addressing the semantic influence brought about by each utterance. Besides, we design an automatic reference-free evaluation metric \textbf{Flow score} based on the pre-trained DialoFlow for interactive dialogue quality evaluation.
    \item The experimental results illustrate that DialoFlow achieves significant improvements on dialogue generation compared to the DialoGPT, and Flow score shows impressively high chatbot-level correlation ($r=0.9$) with human ratings.
\end{itemize}

\section{Method}
The proposed DialoFlow models the dynamic information flow in the whole dialogue history by addressing the semantic influence brought about by each utterance in sequence. 

\begin{figure*}[t]
    \centering
    \includegraphics[width=0.9 \textwidth]{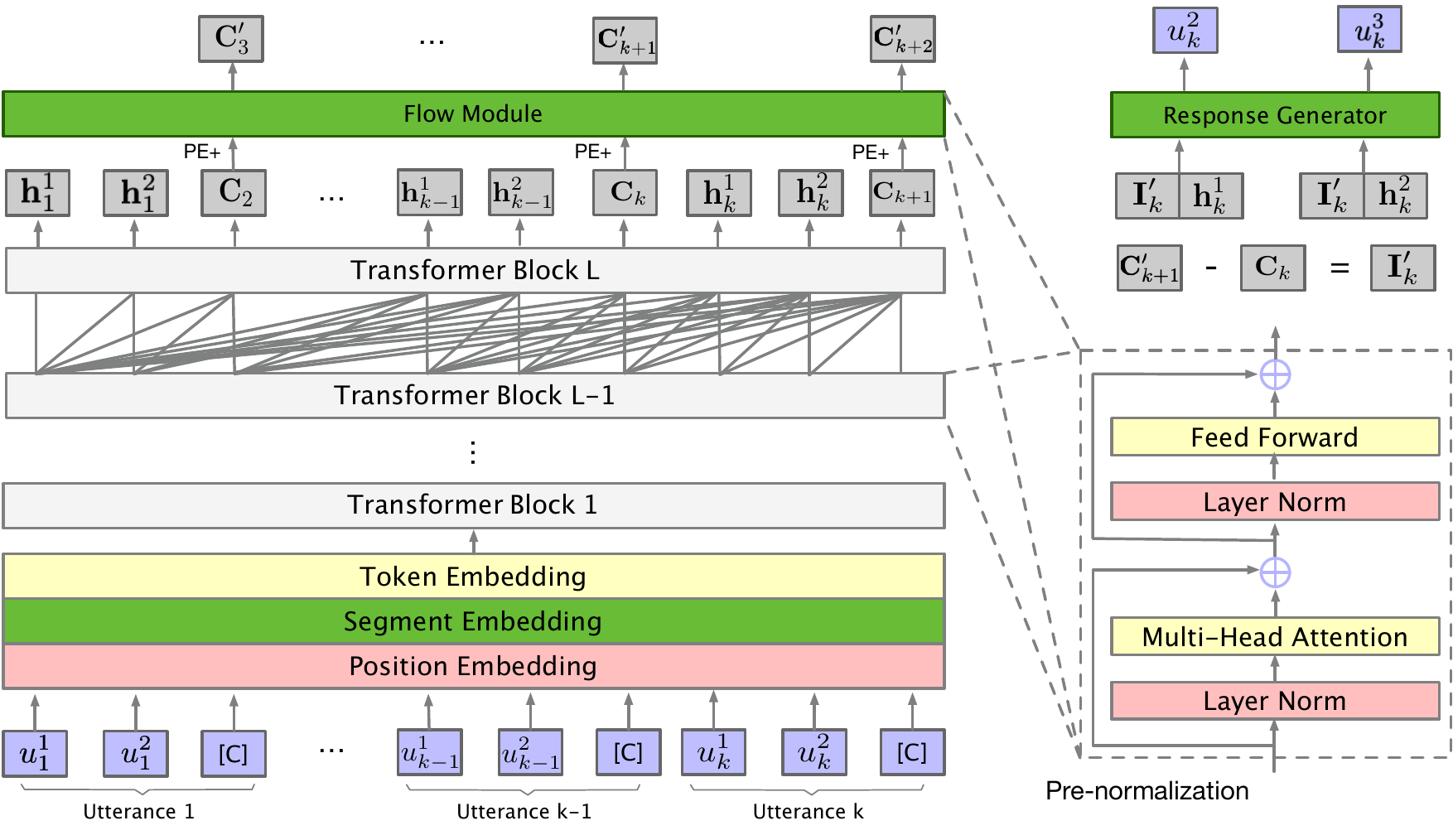}
    \caption{Overview of our DialoFlow. We present the detail model architecture, and the self-attention visualization in DialoFlow. “[C]” is a special token placed at the end of each utterance to model the dense representation of dialogue history $\textbf{C}_k$ named as the context. The future context $\textbf{C}_{k+1}'$ can be predicted by context history $[\textbf{C}_1, \ldots, \textbf{C}_{k}]$. For the simplicity, we only plot two tokens for each utterance.}
    \label{fig:dialoflow}
\end{figure*}

\subsection{Model Overview}

Before introducing the DialoFlow in detail,  we first define some terms. Formally, let $\mathcal{D} = \{u_1, u_2, ..., u_N\}$ denotes a whole dialogue. And for each utterance $u_k = \{u_k^1, u_k^2, ..., u_k^T\}$ where $u_k^t$ denotes the $t$-th word in the $k$-th utterance. We further denote $u_{\textless k} = \{u_1, u_2, ..., u_{k-1}\}$ as the dialogue history at the $k$-th utterance. 
Besides, the dense representation of the dialogue history $u_{<k}$ at the $k$-th utterance is represented as the \textbf{context} $\mathbf{C}_{k}$. 
And the difference between the new context $\mathbf{C}_{k+1}$ at the ($k$+1)-th utterance and the previous contexts $\mathbf{C}_{k}$ at the $k$-th utterance can be defined as the \textbf{semantic influence} $\mathbf{I}_k$ of the $k$-th utterance, which can be formulated as:
\begin{equation}
   \mathbf{I}_k = \mathbf{C}_{k+1} - \mathbf{C}_{k}.
\end{equation}

In our method, DialoFlow first encodes the dialogue history and predicts the future context $\mathbf{C}_{k+1}'$ according to all the previous history context $\mathbf{C}_1, \mathbf{C}_2, ..., \mathbf{C}_{k}$. 
Then at the response generation stage, the model acquires the predicted target semantic influence $\mathbf{I}_k'$, and generate the target response $u_k$ auto-regressively considering both the predicted semantic influence and the historical sub-sentences.
Specifically, as shown in Figure \ref{fig:dialoflow}, DialoFlow models the context flow by designing a uni-directional Flow module upon the transformer, and we introduce three multi-task training objectives to supervise the context flow, semantic influence, and response generation, which are referred to as \emph{context flow modeling}, \emph{semantic influence modeling}, and \emph{response generation modeling}, respectively.

\subsection{Model Architecture}
Figure \ref{fig:dialoflow} demonstrates the infrastructure of DialoFlow, which consists of the input embeddings, transformer blocks, a uni-directional Flow module, and a response generator. 

\noindent \textbf{Input Embedding.} DialoFlow takes the sum of token embedding, segment embedding, and position embedding as the model input. In particular, we insert a special token “[C]” at the end of each utterance, which is used to capture the overall dense representation of the dialogue history. To enhance the modeling of different speakers, we utilize segment embedding containing two types: “[Speaker1]” and “[Speaker2]”.

\noindent \textbf{Transformer Block.} A transformer block consists of the following key components: layer normalization, multi-head attention, and feed-forward layers. We employ the pre-normalization used in GPT-2 \cite{radford2019language} instead of the post-normalization used in BERT \cite{devlin2019bert}, as \citep{shoeybi2019megatron} show that the post-normalization leads to performance degradation when the model size increases while pre-normalization enables stable large-scale training. 
DialoFlow keeps the uni-directional dialogue encoding and enables training on the dialogue level rather than on the context-response setting.
We can obtain the history context at the $k$-th utterance encoded by the transformer blocks:
\begin{equation}
    \textbf{C}_{k} = \text{Transformer}(u_{<k}),
\end{equation}
where $\mathbf{C}_k$ is the hidden states at the position of special token “[C]”.
And the hidden states at the position of each token $u_k^t$ in the input sequence are denoted as $\mathbf{h}_k^t$.

\noindent \textbf{Flow Module.} To capture the dynamic information flow across the dialogue utterances, we design a Flow module to model the context changing scheme. The architecture of the Flow module is the same with one layer of transformer block. The Flow module takes all the previous context $\{\mathbf{C}_1, \mathbf{C}_2, ..., \mathbf{C}_{k}\}$ as input and predicts the context at the ($k$+1)-th utterance $ \mathbf{C}_{k+1}'$:
\begin{equation}
    \mathbf{C}_{k+1}' = \mathrm{Flow}(\mathbf{C}_1, \mathbf{C}_2, ..., \mathbf{C}_{k}).
\end{equation}
The predicted semantic influence brought about by the $k$-th utterance can be computed as:
\begin{equation}
    \mathbf{I}_{k}' = \mathbf{C}_{k+1}' - \mathbf{C}_{k}.
\end{equation}

\noindent \textbf{Response Generator.} DialoFlow generates the utterance $u_k$ with the guidance of the predicted semantic influence $\mathbf{I}_k'$.  The response generator contains a feed-forward layer and a softmax layer to convert the hidden states to tokens. When generating the $t$-th word, the response generator takes the predicted semantic influence $\mathbf{I}_k'$ and the hidden states $\mathbf{h}_k^{t-1}$ as input, and outputs the probability distribution of the $t$-th word:
\begin{align}
     p(u_{k}^{t}|&\mathbf{I}_k', u_{\textless k}, u_k^{\textless t}) = \nonumber \\ &\mathrm{softmax}(W_1[\mathbf{I}_k';\mathbf{h}_k^{t-1}] + b_1) \in \mathbb{R}^{|V|},
\end{align}
where $|V|$ refers to the vocabulary size, $W_1$ and $b_1$ are learnable parameters.

\subsection{Training Objectives}
Different from traditional training approaches with context-response pair, DialoFlow is trained with the whole dialogue containing $N$ utterances. Correspondingly, we design three training tasks to optimize the model: 1) Context Flow Modeling, 2) Semantic Influence Modeling, and 3) Response Generation Modeling.

\noindent \textbf{Context Flow Modeling.}
To capture the dynamic context flow, DialoFlow predicts the context at the $k$-th utterance $\mathbf{C}_k'$ based on the previous context sequence $\{\mathbf{C}_1, ..., \mathbf{C}_{k-1}\}$. We minimize the L2 distance between the predicted context $\mathbf{C}_k'$ and the real context $\mathbf{C}_k$:
\begin{equation}
    \mathcal{L}_{CFM} = \sum_{k=1}^{N}||\mathbf{C}_k - \mathbf{C}_k'||_2^2.
\end{equation}

\noindent \textbf{Semantic Influence Modeling.} To force the effectively modeling of  semantic influence brought about by the $n$-th utterance at the context $\mathbf{C}_{n-1}$, we design a bag-of-words loss using the predicted semantic influence $\mathbf{I}_n'$:
\begin{align}
    \mathcal{L}_{SIM} &= -\sum_{k=1}^{N}\sum_{t=1}^{T} \log p(u_{k}^{t}|\mathbf{I}_k') \nonumber \\ 
                      &= -\sum_{k=1}^{N}\sum_{t=1}^{T} \log f_{u_{k}^{t}},
\end{align}
where $f_{u_{k}^t}$ denotes the estimated probability of the $t$-th word $u_{k}^{t}$ in the utterance $u_k$. The function $f$ is used to predict the words in the utterance $u_k$ in a non-autoregressive way:
\begin{equation}
    f = \mathrm{softmax}(W_2\mathbf{I}_k' + b_2) \in \mathbb{R}^{|V|},
\end{equation}
where $|V|$ refers to the vocabulary size, $W_2$ and $b_2$ are learnable parameters.

\noindent \textbf{Response Generation Modeling.} The predicted semantic influence $\mathbf{I}_k'$ can also be regarded as a semantic expectation of the $k$-th utterance. We incorporate the predicted semantic influence $\mathbf{I}_k'$ into the response generation stage to guide the generation. The response generation objective is as follows:
\begin{align}
    \mathcal{L}_{RGM} &= -\sum_{k=1}^{N} \log p(u_{k}|\mathbf{I}_k', u_{\textless k}) \nonumber \\
                      &= -\sum_{k=1}^{N}\sum_{t=1}^{T} \log p(u_{k}^{t}|\mathbf{I}_k', u_{\textless k}, u_k^{\textless t}).
\end{align}

The overall training objective of DialoFlow can be computed as follows:
\begin{equation}
	\mathcal L = \mathcal L_{CFM} + \mathcal L_{SIM} + \mathcal L_{RGM}.
\end{equation}

\subsection{Flow Score}
By optimizing with the aforementioned three training objectives, DialoFlow can capture the dynamic information flow across the dialogue history. 
As the DialoFlow is trained on human-human dialogues, the context flow scheme can be regarded as the general expectation of the dialogue development. 
Therefore, the closer gap between the semantic influence brought by the chatbot’s utterance and the expectation means the more human-likeness. 

Based on the consideration, we propose an automatic reference-free metric \textbf{Flow score} for interactive dialogue evaluation based on DialoFlow. 
In the human-bot conversation, when the bot generates a new utterance $u_k$, we measure the similarity between the predicted semantic influence $\mathbf{I}_k'$ and the real semantic influence $\mathbf{I}_k$ brought about by the utterance $u_k$, which can be considered as the probability of the human-likeness of the utterance. 
To compute the similarity between the semantic influences, we measure both the cosine similarity and the length similarity:
\begin{align}
    s_k &= \cos(\langle \mathbf{I}_k', \mathbf{I}_k \rangle) \cdot length(\mathbf{I}_k', \mathbf{I}_k) \nonumber \\
        &= \frac{\mathbf{I}_k' \cdot \mathbf{I}_k}{||\mathbf{I}_k'||~||\mathbf{I}_k||} \cdot \frac{\min(||\mathbf{I}_k'||, ||\mathbf{I}_k||)}{\max(||\mathbf{I}_k'||, ||\mathbf{I}_k||)}.
\end{align} 
Note that we introduce the length similarity to consider the influence of length difference on semantic similarity.
For the overall quality of the chatbot in the dialogue, we design a metric, which can be regarded as the dialogue-level perplexity:
\begin{equation}
    Flow~ score = 2 ^{-\frac{1}{M}\sum_{k}^M \log (\frac{s_k+1}{2})},
\end{equation}
where $M$ denotes the turn numbers of the chatbot utterances and $\frac{s_k+1}{2}$ is to scale the similarity value to $[0,1]$. 
A lower Flow score corresponds to better dialogue quality.

\section{Experiments}

\subsection{Dataset}
\noindent \textbf{For model pre-training}, we use the Reddit comments, which are collected by a third party and made publicly available on pushshift.io \cite{baumgartner2020pushshift}. We clean the data following the pipeline used in the DialoGPT.\footnote{https://github.com/microsoft/DialoGPT}

\noindent \textbf{For response generation}, we employ the multi-reference Reddit Test Dataset \cite{zhang2019dialogpt} which contains 6k examples with multiple references. We evaluate our pre-trained DialoFlow model on this dataset. The average length of the dialogue history in this dataset is 1.47. To further explore the dynamic information flow in the long dialogue history situation, we choose another popular open-domain dialogue dataset -- DailyDialog Dataset \cite{li2017dailydialog}, in which the average dialogue history length is about 4.66. DialoFlow is fine-tuned on the DailyDialog training set and evaluated on the DailyDialog multi-reference test set \cite{DBLP:conf/sigdial/GuptaMZPEB19}.

\noindent \textbf{For interactive dialogue quality evaluation}, we employ the collected data from the Interactive Evaluation of Dialog Track @ The Ninth Dialog System Technology Challenge (DSTC9) \cite{gunasekara2020overview}, which contains 2200 human-bot conversations from 11 chatbots. For each conversation, there are 3 human ratings on the overall quality (0-5).
We calculate the correlation between the results of our proposed metric and the human ratings on the chatbot level.
Human-human conversations are always regarded to be better than human-bot conversations. Therefore, we randomly sample 200 human-human dialogues from the BST \cite{smith2020can} dataset to see the metric’s performance on the real human-human conversations.

\subsection{Experimental Setting}

\noindent \textbf{Pre-training Details.} DialoFlow is pre-trained based on the pre-trained GPT-2 \cite{radford2019language}, since \citet{zhang2019dialogpt} show that DialoGPT trained from the pre-trained GPT-2 is much better than from scratch. There are three different model sizes: DialoFlow-base, DialoFlow-medium, and DialoFlow-large, which are trained from the pre-trained GPT2-base, GPT2-medium, GPT2-large, respectively. We used AdamW optimizer \cite{DBLP:conf/iclr/LoshchilovH19} with 0.01 weight decay and linear learning rate scheduler with 12000 warm-up steps. The learning rate is 2e-4 for the base and medium version and 1e-4 for the large version. We use the batch size of 1024 for all model sizes. 
We trained the base and medium models for up to 4 epochs and trained the large model for 2 epochs. It costs about two months on 8 Nvidia V100 GPUs to train the large model.

\noindent \textbf{Decoding Details.} On the 6K Reddit multi-reference dataset, we use beam search (with beam width 10) on the DialoFlow-medium model and the DialoFlow-large model. We employ greedy search on the DialoFlow-base model, which keeps the same with \citep{zhang2019dialogpt}. On the DailyDialog dataset, we fine-tune the pre-trained DialoFlow and DialoGPT, select the checkpoint based on the validation loss, and then use beam search (with beam width 5) for decoding.

\subsection{Baseline}
\textbf{For response generation}, we compare our proposed DialoFlow with DialoGPT, a popular dialogue generation model pre-trained on the Reddit Comments. We choose the version trained from pre-trained OpenAI GPT-2 for comparison. 

\noindent \textbf{For interactive dialogue evaluation}, we compare our metric with the following metrics:
    1) \textbf{FED score} \cite{DBLP:conf/sigdial/MehriE20} is an automatic evaluation metric which uses DialoGPT-large, without any fine-tuning or supervision. FED takes the DialoGPT-large as the user and calculates the likelihood of follow-up utterances based on several pre-set usual human utterances. FED works under the pre-set common human utterances, which can reveal the dialogue quality.
    2) \textbf{Perplexity} is used to measure the coherence of an utterance under the dialogue context. We employ DialoGPT-large to measure the perplexity for each utterance of the chatbot. We average the perplexity of all utterances in the whole dialogue as the baseline metric.

\begin{table*}[h]
		\centering
		\begin{tabular}{l|cc|cc|c|c|c}
			\toprule[1pt]
			\bf{Method} & NIST-2 & NIST-4 & BLEU-2 & BLEU-4 & METEOR & Entropy & Avg Len \\
			\hline
			\multicolumn{8}{c}{Multi-reference Reddit Dataset} \\
			\hline
			DialoGPT (B, greedy) & 2.39  & 2.41 & 10.54\%  & 1.55\% & 7.53\% & \bf 10.77  & 12.82  \\
			DialoFlow (B, greedy) & 2.88  & 2.93 & 15.34\%  & 3.97\% & 9.52\% & 9.27  & \bf 15.43  \\
			DialoGPT (M, beam) & 3.40  & 3.50 & \bf 21.76\%  & \bf 7.92\% & 10.74\% & 10.48 & 11.34  \\
			DialoFlow (M, beam) & 3.89  & 3.99 & 20.98\%  & 7.36\% & 11.46\% & 10.42 & 13.37  \\
			DialoGPT (L, beam) & 2.90  & 2.98 & 21.08\%  & 7.57\% & 10.11\% & 10.06 & 10.68  \\
			DialoFlow (L, beam) & \bf 3.90  & \bf 4.01 &  21.20\%  & 7.42\% & \bf 11.48\% & 10.42 & 13.38  \\
			\hline
			Human & 3.41  & 3.50 & 17.90\%  & 7.48\% & 10.64\% & 10.99 & 13.10  \\
			\midrule[1pt]
			\multicolumn{8}{c}{Multi-reference DailyDialog Dataset} \\
			\hline
			DialoGPT (B, beam) & 2.28  & 2.78 & 18.83\%  & 6.63\% & 15.5\% & 9.80 & \bf{18.82}  \\
			DialoFlow (B, beam) & 3.65  & 3.84 & 26.47\%  & 10.12\% & 16.1\% & 9.62 & 12.00  \\
			DialoGPT (M, beam) & 3.47  & 3.65 & 25.39\%  & 9.99\% & 15.9\% & 9.64 & 12.88  \\
			DialoFlow (M, beam) & 3.80  & 4.02 &  27.63\%  &  11.33\% &  16.7\% &  9.83  & 12.06  \\
			DialoGPT (L, beam) & 3.30  & 3.46 & 23.69\%  & 9.20\% & 15.7\% & 9.78 &  13.24  \\
			DialoFlow (L, beam) & \bf 3.86  & \bf 4.08 & \bf 28.02\%  & \bf 11.57\% & \bf 17.0\% & \bf 9.87 & 12.08  \\
			\midrule[1pt]
			\multicolumn{8}{c}{Ablation Study on Multi-reference Reddit Dataset} \\
			\hline
			DialoFlow (M, beam) & 3.89  & 3.99 & 20.98\%  & 7.36\% & 11.46\% & 10.42 & 13.37  \\
			~~ w/o SIM  & 3.85  & 3.96 &  21.36\%  &  7.71\% &  11.26\% &  10.43  & 12.70  \\
			~~ w/o SIM \& CFM & 3.79  & 3.89 & 21.33\%  &  7.65\% &  11.25\% &  10.33  & 12.55  \\
			\bottomrule[1pt]
		\end{tabular}
		\caption{The evaluation on 6K Reddit multi-reference dataset and on DailyDialog dataset. For 6K Reddit multi-reference dataset, as the DialoGPT do not release the decoding code, we directly quote the results from \cite{zhang2019dialogpt}. Note that “B”, “M”, “L” denotes base, medium, large respectively. }
		\label{tab:response_generation}
\end{table*}

\subsection{Evaluation Metrics}
\textbf{For dialogue response generation}, we perform automatic evaluation using common reference-based metrics:
BLEU \cite{DBLP:conf/acl/PapineniRWZ02}, METEOR \cite{DBLP:conf/wmt/LavieA07}, and NIST \cite{lin2004automatic}. NIST is a variant of BLEU that weights $n$-gram matches by their information gain, i.e., it indirectly penalizes uninformative $n$-grams such as “I  don’t know”, which is a more suitable metric than BLEU when dealing with multi-reference test sets. We also use Entropy \cite{DBLP:conf/nips/ZhangGGGLBD18} to evaluate the lexical diversity. We employ the evaluation scripts used by DialoGPT.

\noindent \textbf{For interactive dialogue evaluation}, we compute the Pearson and Spearman correlation between the automatic metrics and human ratings. We use the pre-trained DialoFlow-large to compute our proposed Flow score. 

\begin{table}[t]
    \small
		\centering
		\begin{tabular}{c|c|c|c}
			\toprule[1pt]
			\bf{Metric} & DialoFlow & DialoGPT & Tie \\
			\hline
			\bf{Relevance} & 43.7\% & 28.8\% & 27.5\% \\
			\bf{Informativeness} & 45.3\% & 29.2\% & 25.5\%\\
			\bf{Human-likeness} & 46.2\% & 29.3\% & 24.5\%\\
			\bottomrule[1pt]
		\end{tabular}
		\caption{Human evaluation for DialoFlow and DialoGPT on the DailyDialog test Dataset.}
		\label{tab:human}
\end{table}

\begin{table*}[t]
		\centering
		\small
		\begin{tabular}{c|ccccccccccc|c}
			\toprule[1pt]
			 \bf Methods & \bf B1 & \bf B2 & \bf B3 & \bf B4 & \bf B5 & \bf B6 & \bf B7 & \bf B8 & \bf B9 & \bf B10 & \bf B11 & \bf Human  \\
			\hline
			\bf{Human $\uparrow$} & 4.142 & 4.140 & 4.075 & 4.035 & 3.933 & 3.864 & 3.849 & 3.848 & 3.828 & 3.692 & 3.605 & \emph{5.000} \\
			\bf{FED $\uparrow$} & 4.988 & 4.818 & 4.621 & 4.670 & 4.555 & 4.739 & 4.438 & 4.355 & 4.651 & 4.799 & 3.608  & 3.468 \\
			\bf{Perplexity $\downarrow$} & 600.0 & 521.2 & 441.2 & 561.6 & 367.7 & 1731 & 1879 & 13347 & 662.2 & 618.4 & 50.29 & 51.39 \\
			\bf{Flow $\downarrow$} & 1.396 & 1.410 & 1.402 & 1.406 & 1.407 & 1.422 & 1.425 & 1.417 & 1.425 & 1.461 & 1.466 & 1.333 \\
			\bottomrule[1pt]
		\end{tabular}
		\caption{The human ratings and automatic metrics for different chatbots. B1$\sim$B11 denotes the 11 different chatbots in the DSTC9 Interactive Dialogue Evaluation Track. Human denotes the performance on the human-human conversations from the BST dataset. We assume the human rating for the human-human conversations is 5.000.}
		\label{tab:flow_score}
\end{table*}

\begin{table}[t]
        \small
		\centering
		\begin{tabular}{c|c|c}
			\toprule[1pt]
			\bf{Method} & Pearson & Spearman \\
			\hline
			\bf{FED} & 0.67 (p \textless 0.1) & 0.56 (p \textless 0.1) \\
			\bf{Perplexity} & 0.12 (p$\approx$0.72) & 0.20 (p$\approx$0.55) \\
			\bf{Flow} & \textbf{0.91} (p \textless 0.001) & \textbf{0.90} (p \textless 0.001) \\
			\bottomrule[1pt]
		\end{tabular}
		\caption{Chatbot-level correlations on the DSTC9 Interactive Conversation dataset.}
		\label{tab:correlation}
\end{table}

\section{Results and Analysis}
In this section, we show the performance of our pre-trained DialoFlow model on response generation as well as the performance of Flow score on interactive dialogue quality evaluation.
\subsection{Response Generation}
\begin{figure}[t]
    \centering
    \includegraphics[width=0.45 \textwidth]{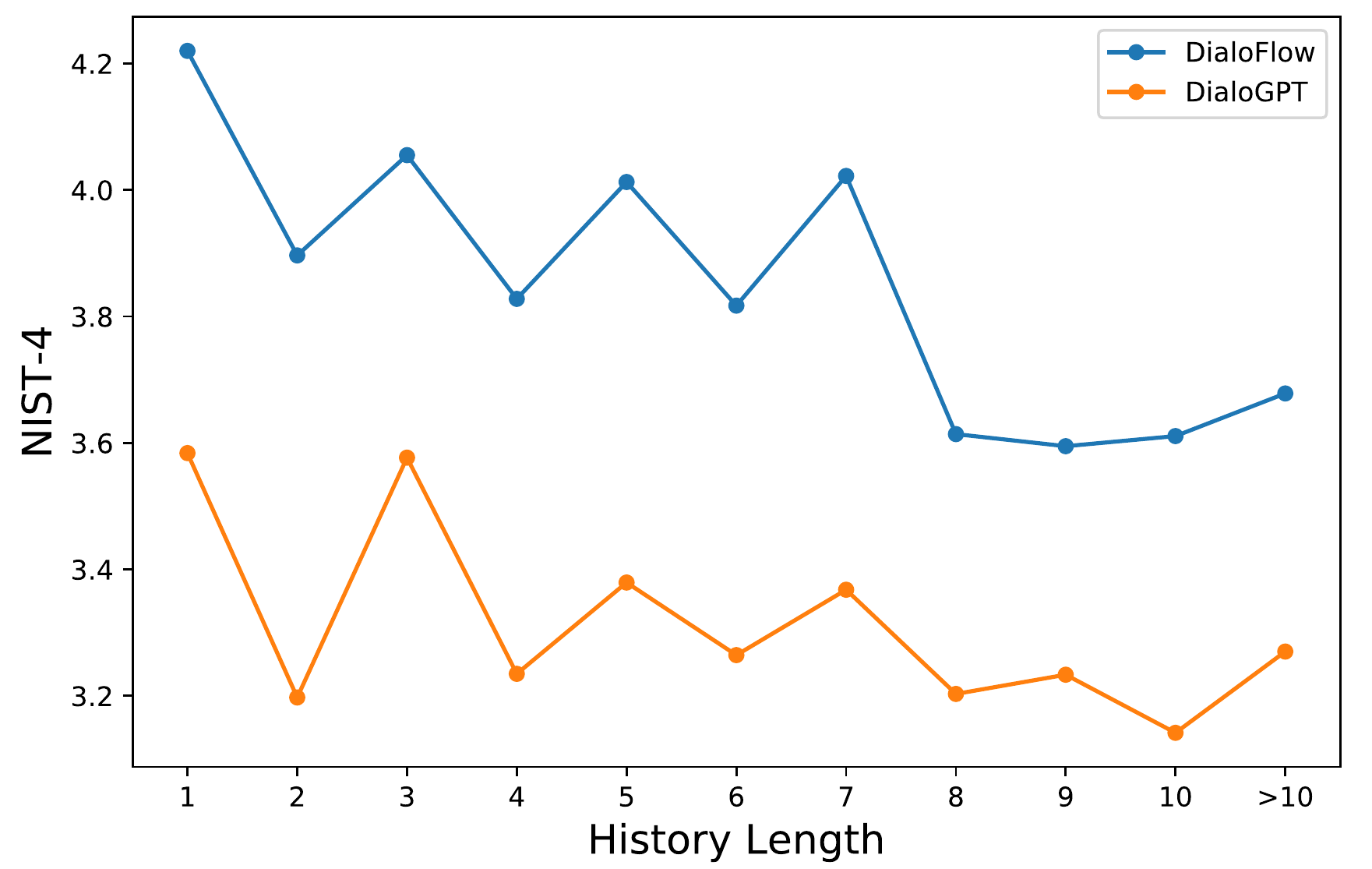}
    \caption{The performance of the large version of DialoFlow and DialoGPT on the samples of different history lengths on the DailyDialog dataset.}
    \label{fig:length}
\end{figure}

Table \ref{tab:response_generation} lists the comparison of our pre-trained DialoFlow with the pre-trained DialoGPT on the Reddit multi-reference dataset. Generally, DialoFlow-large achieves the highest score on the NIST and METEOR, while DialoGPT-medium performs better on the BLEU. 
The performance of our DialoFlow increases with the model size, while the DialoGPT gets the best performance with the medium size rather than the large size.
As NIST can effectively penalize common n-grams such as “I don’t know”, the results reveal that DialoGPT tends to generate general responses while our DialoFlow model can create more informative responses. 
The results also reflect that modeling the dynamic flow is helpful to boost the conversion quality and avoid converging to the general responses. 
For the lexical diversity, DialoFlow performs similarly with the DialoGPT on Entropy. 

The average history length of the multi-reference Reddit dataset is only 1.45, which is a bit short.  Thus, we conduct extensive experiments on the DailyDialog dataset (average history length = 4.66) to verify the performance gain on the long dialogue history. As shown in Table \ref{tab:response_generation}, DialoFlow shows significant improvements on all model sizes and on all metrics compared to the DialoGPT. The improvements on the DailyDialog dataset demonstrate that our DialoFlow model shows a great capacity to capture the dynamic information flow with a long history. 
Note that the performance improvement of the DailyDialog dataset is more remarkable than Reddit. In our opinion, conversations in Reddit are mainly the comments in forums, while in DailyDialog the dialogues are derived from daily life. Thus, in the DailyDialog dataset, the context flows are in the more similar schema, and the semantic influences are more predictable compared to the Reddit dataset.   

\noindent \textbf{Human Evaluation.} We conduct human evaluation on 200 randomly sampled cases from the DailyDialog test dataset using crowd-sourcing. We compare DialoFlow and DialoGPT on the medium version. Each response pair is randomly presented to 3 judges, who rank them for relevance, informativeness, and human-likeness. The overall judge preferences are presented as a percentage of the total, as shown in Table \ref{tab:human}. There is a strong preference for the responses generated by DialoFlow. The human evaluation demonstrates that modeling the dynamic information flow is effective for improving the quality of dialogue generation.


\noindent \textbf{Analysis of dialogue history length.} 
Figure \ref{fig:length} shows the performance of our DialoFlow and the DialoGPT on different history lengths. Overall, our DialoFlow achieves better performance on all history lengths. In particular, when history length equals 1, that is, the response is generated based on one history utterance, our DialoFlow also gains a prominent boosting. We attribute it to the guidance of predicted semantic inference.

\noindent \textbf{Ablation Study.}
To explore the effect of the proposed training objectives, we conduct ablation studies on the medium version of DialoFlow, as shown in Table \ref{tab:response_generation}. With all three training objectives, DialoFlow model achives the best performance on NIST and METEOR. When we drop the Semantic Influence Modeling task, the performance slightly decreases. When we further drop the Context Flow Modeling task, which means the end-to-end training, the performance decreases again. The results reveal that the Context Influence Modeling task is effective for dialogue modeling and the Semantic Influence Modeling task can prompt the CIM task.

\begin{figure*}[t]
    \centering
    \includegraphics[width=0.95 \textwidth]{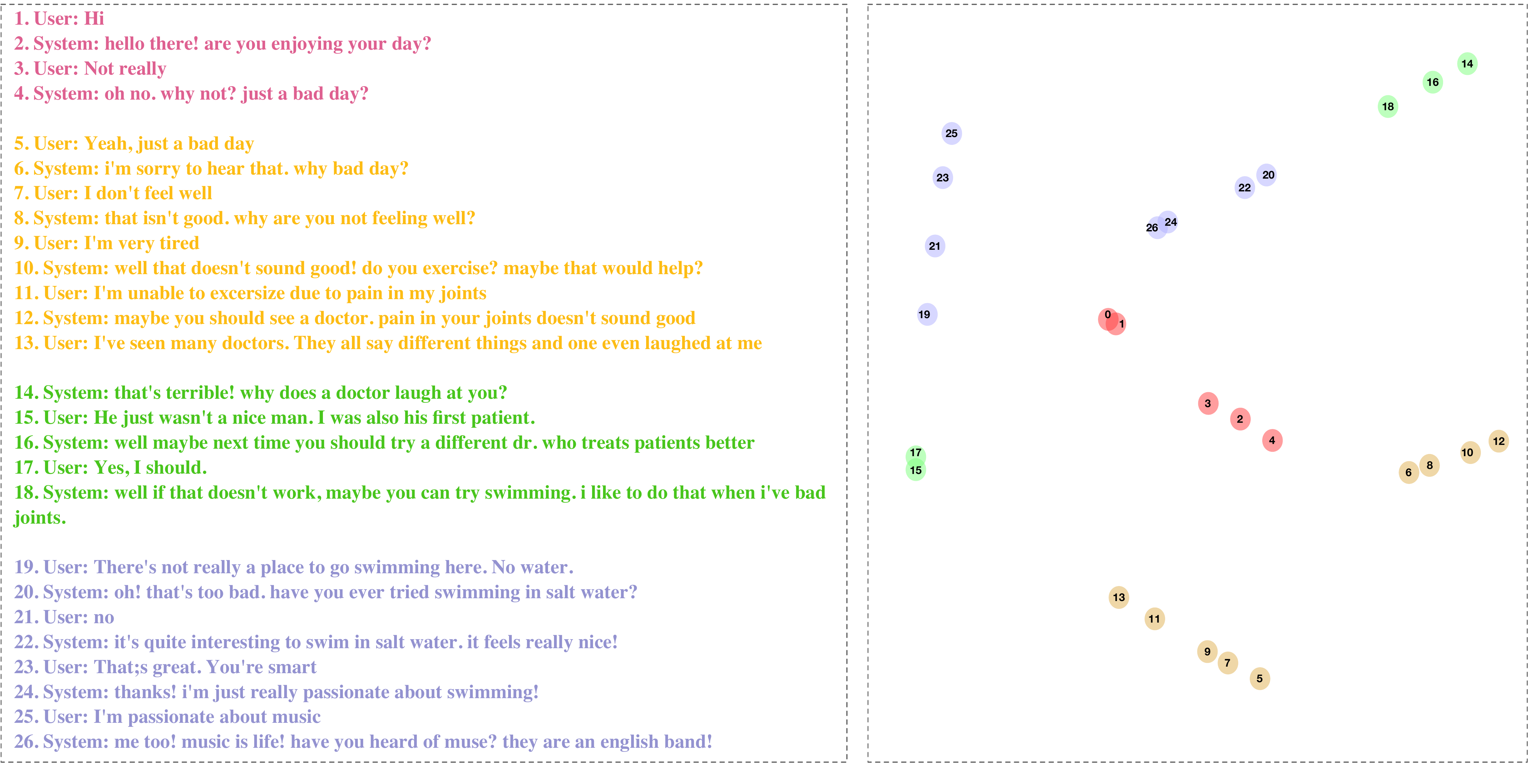}
    \caption{The semantic context 2-D T-SNE visualization of a human-bot conversation.  Our DialoFlow model captures the context flow, especially the topic changes. “0” is the start point. Better view in color.}
    \label{fig:case_study}
\end{figure*}

\subsection{Dialogue Evaluation}
\noindent \textbf{Results.} Table \ref{tab:correlation} shows the chatbot-level correlations of different automatic metrics with human ratings on the DSTC9 Interactive Conversation dataset. Our proposed Flow score achieves strong Spearman correlation of 0.90 ($p\textless0.001$) and strong Pearson correlation of 0.91 ($p\textless0.001$). FED only shows moderate correlations with a chatbot-level Spearman correlation of 0.56 ($p\textless0.1$). Perplexity score shows a very weak correlation. On the one hand, the results reveal that our proposed Flow score can effectively estimate the overall chatbot quality.  On the other hand, high correlation also demonstrates that the DialoFlow model captures the general dynamic information flow in the natural human-human conversation.

\noindent \textbf{Results Analysis.} Table \ref{tab:flow_score} shows the detailed human ratings, FED scores, perplexity, and our proposed Flow score for the 11 chatbots in the DSTC9 Interactive Dialogue Evaluation Track and the sampled human-human conversations. Good automatic metrics should perform well not only on human-bot conversations but also human-human conversations because the ultimate goal of the chatbot is to generate human-like responses. FED performs poorly on the human-human conversations compared to its performance on the other 11 chatbots. Our proposed Flow score takes the human-human conversations as the best one, and the Flow score gap between human-human conversations and the best chatbot is similar to the human rating gap.

\noindent \textbf{Analysis about Flow score.} The Flow score can be regarded as the perplexity on the utterance level. There are many different expressions for a specific semantic in natural conversations. Traditional word-level perplexity can estimate the coherence and fluency of the utterance but always performs unstably on variable expressions. The Flow score directly measures the semantic similarity and alleviates the problem with the traditional perplexity.

\subsection{Case Study}

Figure \ref{fig:case_study} shows the 2-D T-SNE visualization of the semantic context of a human-bot conversation encoded by our pre-trained DialoFlow model. The conversation can be split into four topics: greetings (1$\sim$4), talking about why bad day (5$\sim$13), explaining the terrible experience seeing the doctor (14$\sim$18), and discuss swimming (19$\sim$26). Correspondingly, in the visualization, the semantic context flow visualization changes a lot when the topic switches, revealing that DialoFlow can capture the dynamic information flow in the dialogue and effectively measure the semantic influence brought about by each utterance. Besides, it seems like that different speakers keep their own context flows.

\section{Related Works}

\noindent \textbf{Multi-turn dialogue modeling.}
The modeling of multi-turn dialogue history mainly falls into two categories: 1) Flat concatenation. These works directly concatenate the dialogue history as the input sequence \cite{zhang2019dialogpt}, which can not capture the information dynamics. 2) Hierarchical architectures. The hierarchical architecture is commonly used in the dialogue history understanding. \citet{serban-etal-2016-generating} propose the hierarchical LSTM to generate responses. \citet{DBLP:conf/acl/LiNMFLZ19} introduce an incremental transformer to capture multi-turn dependencies. \citet{DBLP:conf/acl/ShanLZMFNZ20,gu2020dialogbert} employ pre-trained BERT to encode individual utterances and design the utterance-level encoder to capture the turn-level structure. These methods suffer from the lack of context word-level information when encoding utterances. Different from these methods, our DialoFlow takes full advantage of both word-level information and utterance-level dynamic information. Besides, the proposed DialoFlow is pre-trained on the large-scale open-domain dialogue dataset.
 
\noindent \textbf{Pre-trained models for dialogue generation.}
Recent advances in pre-trained language models have great success in dialogue response generation. DialoGPT\cite{zhang2019dialogpt}, Plato-2 \cite{bao2020plato}, Meena\cite{adiwardana2020towards}, and Blender\cite{smith2020can} achieve strong generation performances by training transformer-based language models on open-domain conversation corpus. In contrast, our proposed DialoFlow focuses on modeling the dynamic information flow in the pre-training process, and we design three training objectives to optimize the model. 

\noindent \textbf{Interactive Dialogue Evaluation.} 
Evaluating the quality of interactive dialogue automatically is a challenging problem, as there is no gold reference for the utterances. 
\citet{DBLP:conf/sigdial/MehriE20} propose the FED score, an automatic dialogue evaluation metric using pre-trained DialoGPT-large, which works with pre-set common human comments, like \emph{“It is interesting to talk with you.”}, revealing the dialogue quality. However, the FED score has limited performance on those dialogues without apparent comments.
Our Flow score entirely depends on the pre-trained DialoFlow model with no need for human integration.

\section{Conclusion and Future work}
In this work, we proposed the DialoFlow to model the dynamic information flow across dialogue utterances by addressing the semantic influence brought about by each utterance. Specifically, we employed a uni-directional Flow module to model the context flow and designed three training objectives to optimize the DialoFlow model. Besides, upon the DialoFlow, we proposed the Flow score, an automatic reference-free evaluation metric for interactive dialogue evaluation, with the pre-trained DialoFlow. Experiments on response generation and dialogue evaluation all demonstrate that our method could effectively capture the dynamic information flow across utterances. 
For future work, we would like to apply the DialoFlow to the task-oriented dialogue and explore the application on the long text generation, such as the story generation.

\section*{Acknowledgement}
We sincerely thank the anonymous reviewers for their thorough reviewing and valuable suggestions. This work is supported by National Key R\&D Program of China (NO. 2018AAA0102502).

\bibliographystyle{acl_natbib}
\bibliography{anthology,acl2021}


\end{document}